\pdfoutput=1
\documentclass[11pt,a4paper]{article}
\usepackage{authblk}
\usepackage[hyperref]{emnlp-ijcnlp-2019}
\usepackage{times}
\usepackage{latexsym}
\usepackage{csvsimple}
\usepackage{amsmath}
\usepackage{multicol}
\usepackage{url}
\usepackage{hhline}
\usepackage{devanagari}
\usepackage{tabularx, booktabs}
\usepackage{pifont}
\usepackage{longtable}

\aclfinalcopy 

\title{Small and Practical BERT Models for Sequence Labeling}

\author{Henry Tsai}
\author{Jason Riesa}
\author{Melvin Johnson}
\author{Naveen Arivazhagan}
\author{Xin Li}
\author{Amelia Archer}
\affil{Google Research \\
San Francisco, CA, USA \\
{\tt \{henrytsai,riesa,melvinp,navari,xinl,ameliaarcher\}@google.com}}

\date{}

\newcommand{\bert}{{BERT}}
\newcommand{\metalstm}{{Meta-LSTM}}
\newcommand{\smallbert}{{MiniBERT}}

\newcommand{\mbert}{\textit{m}BERT}
\newcommand{\mmetalstm}{\textit{m}Meta-LSTM}
\newcommand{\msmallbert}{\textit{m}MiniBERT}

\begin{document}
\maketitle
\begin{abstract}
We propose a practical scheme to train a single multilingual sequence labeling model that yields state of the art results and is small and fast enough to run on a single CPU. Starting from a public multilingual BERT checkpoint, our final model is 6x smaller and 27x faster, and has higher accuracy than a state-of-the-art multilingual baseline. We show that our model especially outperforms on low-resource languages, and works on codemixed input text without being explicitly trained on codemixed examples. We showcase the effectiveness of our method by reporting on part-of-speech tagging and morphological prediction on 70 treebanks and 48 languages.
\end{abstract}

\section{Introduction}

There have been many recent modeling improvements \cite{smith-etal-2018-82, bohnet-etal-2018-morphosyntactic} on morphosyntactic tagging tasks. However, these models have largely focused on building separate models for each language or for a small group of related languages. In this paper, we consider the implications of training, evaluating, and deploying a single multilingual model for a diverse set of almost 50 languages, evaluating on both part-of-speech tagging and morphological attribute prediction data from the Universal Dependencies repository~\cite{UD2.2}.

There are several benefits of using one multilingual model over several language specific models.
\begin{itemize}
\setlength\itemsep{0.01em}
   \item Parameter sharing among languages reduces model size and enables cross-lingual transfer learning. We show that this improves accuracy, especially for low-resource languages.
    \item No language identification model needed to decide which language-specific model to query. Critically, this reduces system complexity and prevents prediction errors from language identification from propagating into the downstream system.
    \item Multilingual models can be applied to multilingual or codemixed inputs without explicitly being trained on codemixed labeled examples. Otherwise, given, e.g. a mixed Hindi/English input, one must decide to query either the Hindi model or the English model, both of which are sub-optimal.
\end{itemize}

In this paper, we show that by finetuning a pretrained BERT~\cite{devlin2019} model we can build a multilingual model that has comparable or better accuracy to state-of-the-art language-specific models, and outperforms the state-of-the-art on low-resource languages. Our model outperforms other multilingual model baselines by a large margin. We evaluate on both part-of-speech tagging and morphological attribute prediction tasks with data from the Universal Dependencies repository~\cite{UD2.2}. However, this model is slow, very large, and difficult to deploy in practice.

We describe our solution for making this model small and practical enough to use in practice on a single CPU, while preserving quality. The final model is 27x faster than a BERT-based baseline model and 7x faster than a state-of-the-art LSTM-based model on CPU. It is 6 times smaller than the BERT-based model. Furthermore, most of the quality gains are preserved in the small model.

\section{Multilingual Models for Sequence Labeling}

\begin{table*}[t]
   \small
    \centering
    \begin{tabular}{lccc}
    \toprule
    Model  & Multilingual? & Part-of-Speech F1 & Morphology F1 
    \\\midrule
    \metalstm{}& No       & 94.5          & 92.5 \\
    \bert{}    & No       & \textbf{95.1} & \textbf{93.0}
    \\\midrule
    \metalstm{} & Yes    & 91.1 & 82.9  \\
    \bert{}     & Yes    & \textbf{94.5} & \textbf{91.0} \\
    \bottomrule
    \end{tabular}
    \caption{Macro-averaged F1 comparison of per-language models and multilingual models over 48 languages. For non-multilingual models, F1 is the average over each per-language model trained.} 
    \label{tab:big_mutlilingual_summary}
\end{table*}

We discuss two core models for addressing sequence labeling problems and describe, for each, training them in a single-model multilingual setting: (1) the Meta-LSTM~\cite{bohnet-etal-2018-morphosyntactic}, an extremely strong baseline for our tasks, and (2) a multilingual BERT-based model~\cite{devlin2019}.

\subsection{Meta-LSTM}
The \metalstm{} is the best-performing model of the CoNLL 2018 Shared Task \cite{smith-etal-2018-82} for universal part-of-speech tagging and morphological features. The model is composed of 3 LSTMs: a character-BiLSTM, a word-BiLSTM and a single joint BiLSTM which takes the output of the character and word-BiLSTMs as input. The entire model structure is referred to as \metalstm{}.

To set up multilingual \metalstm{} training, we take the union of all the word embeddings from the \newcite{bojanowski2017enriching} embeddings model on Wikipedia\footnote{\scriptsize https://fasttext.cc/docs/en/pretrained-vectors.html} in all languages. For out-of-vocabulary words, a special unknown token is used in place of the word.

The model is then trained as usual with cross-entropy loss. The char-BiLSTM and word-biLSTM are first trained independently. And finally we train the entire \metalstm{}.

\subsection{Multilingual BERT}
BERT is a transformer-based model \cite{NIPS2017_7181} pretrained with a masked-LM task on millions of words of text. In this paper our BERT-based experiments make use of the cased multilingual BERT model available on GitHub\footnote{\scriptsize https://github.com/google-research/bert/blob/master/multilingual.md} and pretrained on 104 languages. 

Models fine-tuned on top of BERT models achieve state-of-the-art results on a variety of benchmark and real-world tasks. 


To train a multilingual BERT model for our sequence prediction tasks, we add a softmax layer on top of the the first wordpiece~\cite{schuster2012japanese} of each token\footnote{We experimented with wordpiece-pooling~\cite{wp-pooling} which we found to marginally improve accuracy but at a cost of increasing implementation complexity to maintain.} and finetune on data from all languages combined. During training, we concatenate examples from all treebanks and randomly shuffle the examples.

\section{Small and Practical Models}
The results in Table~\ref{tab:big_mutlilingual_summary} make it clear that the BERT-based model for each task is a solid win over a Meta-LSTM model in both the per-language and multilingual settings. However, the number of parameters of the BERT model is very large (179M parameters), making deploying memory intensive and inference slow: 230ms on an Intel Xeon CPU. Our goal is to produce a model fast enough to run on a single CPU while maintaining the modeling capability of the large model on our tasks.

\begin{table}[t]
\small
    \centering
    \begin{tabular}{lrrrr}
\toprule
Model & Embedding & Tokens & Hidden & Layers \\
      &  Size     &           & Units  &
\\\midrule
\metalstm& 300& 1.2M& 8M& 3\\
\bert& 768& 120k& 87M& 12\\
\smallbert& 256& 120k& 2M& 3 
\\\bottomrule
\end{tabular}
\caption{The number of parameters of each model. \textit{Tokens} refers to the number of tokens of the embedding rows. For the \metalstm{}, a word-based model, this is the number of words in training. For \bert{}, this means the size of the Wordpiece vocabulary. And \textit{Hidden Units} refers to all units that are not among the embedding layer or and output layer.}
\label{tab:modelstructure}
\end{table}

\begin{table}
\centering
\small
\begin{tabular}{lrr}
\toprule
Input & 32 words  & 128 words\\
\midrule
\textit{Relative Speedup on GPU} &&\\
\metalstm{}  &{0.8x}& 0.2x \\
\smallbert  &\textbf{4.3x} & \textbf{2.6x}\\
\midrule
\textit{Relative Speedup on CPU} &&\\
\metalstm{}  & {6.8x}  & 2.3x \\
\smallbert{} & \textbf{27.7x} & \textbf{14.0x} \\
\bottomrule
\end{tabular}
\caption{Relative inference speedup over BERT. We see \smallbert{} is the fastest on both CPU and GPU. CPU is an Intel Xeon CPU E5-1650 v3 @3.50GHz. GPU is an Nvidia Titan~V.}
\label{tab:modelspeed}
\end{table}

\subsection*{Size and speed}
We choose a three-layer BERT, we call \textit{\smallbert{}}, that has the same number of layers as the \metalstm{} and has fewer embedding parameters and hidden units than both models. Table \ref{tab:modelstructure} shows the parameters of each model. The \metalstm{} has the largest number of parameters dominated by the large embeddings. BERT's parameters are mostly in the hidden units. The \smallbert{} has the fewest total parameters.

The inference-speed bottleneck for \metalstm{} is the sequential character-LSTM-unrolling and for BERT is the large feedforward layers and attention computation that has time complexity quadratic to the sequence length.   Table \ref{tab:modelspeed} compares the model speeds.

BERT is much slower than both MetaLSTM and \smallbert{} on CPU. However, it is faster than \metalstm{} on GPU due to the parallel computation of the transformer. The \smallbert{} is significantly faster than the other models on both GPU and CPU.

\subsection*{Distillation}

For model distillation~\cite{hinton2015distilling}, we extract sentences from Wikipedia in languages for which public multilingual is pretrained.
For each sentence, we use the open-source BERT wordpiece tokenizer~\cite{schuster2012japanese,devlin2019} and compute cross-entropy loss for each wordpiece:

    
\[
L(\textbf{t}, \textbf{s}) = H(\sigma(\textbf{t}/T), \sigma(\textbf{s}/T))
\]

\noindent
where $H$ is the cross-entropy function, $\sigma$ is the softmax function, $\textbf{t}$ is the BERT model's logit of the current wordpiece, $\textbf{s}$ is the small BERT model's logits and $T$ is a temperature hyperparameter, explained in Section~\ref{Sec:Tuning}.

To train the distilled multilingual model \msmallbert{}, we first use the distillation loss above to train the student from scratch using the teacher's logits on unlabeled data. Afterwards, we finetune the student model on the labeled data the teacher is trained on.

\section{Experiments}

\subsection{Data}
We use universal part-of-speech tagging and morphology data from the The CoNLL 2018 Shared Task \cite{UD2.2, K18-2:2018}. For comparison simplicity, we remove the languages that the multilingual BERT public checkpoint is not pretrained on.

For segmentation, we use a baseline segmenter (UDPipe v2.2)\footnote{https://ufal.mff.cuni.cz/udpipe/models} provided by the shared task organizer to segment raw text. We train and tune the models on gold-segmented data and apply the segmenter on the raw test of test data before applying our models.

The part-of-speech tagging task has 17 labels for all languages. For morphology, we treat each morphological group as a class and union all classes as a output of 18334 labels.


\subsection{Tuning}
\label{Sec:Tuning}
For \metalstm{}, we use the public repository's hyperparameters\footnote{https://github.com/google/meta\_tagger}.

Following \newcite{devlin2019}, we use a smaller learning rate of 3e-5 for fine-tuning and a larger learning rate of 1e-4 when training from scratch and during distillation. Training batch size is set to 16 for finetuning and 256 for distillation.

For distillation, we try temperatures $T = {1, 2, 3}$ and use the teacher-student accuracy for evaluation. We observe BERT is very confident on its predictions, and using a large temperature $T=3$ to soften the distribution consistently yields the best result.





\subsection{Multilingual Models}

\begin{table}[t]
\small
    \centering
    \begin{tabular}{lccc}
    \toprule
    Model  & Part-of-Speech F1 & Morphology F1
    \\\midrule
    \mmetalstm{} & 91.1 & 82.9  \\
    \msmallbert{} & {93.7} &  88.6\\
    \mbert{} & \textbf{94.5} & \textbf{91.0} \\
    \bottomrule
    \end{tabular}
    \caption{Macro-averaged F1 comparison of multilingual models. Multilingual models are prefixed with `\textit{m}'.} 
    \label{tab:mutlilingual_summary}
\end{table}



\begin{table*}[t]
\small
    \centering
    \begin{tabular}{lcccccc||cccccc}
    \toprule
         &  \multicolumn{6}{c}{POS Tagging}&\multicolumn{6}{c}{Morphology} \\
    Languages  & kk & hy & lt & be & mr&ta & kk & hy & lt & be & mr&ta\\
    \midrule
    Train Size & 31 & 50 & 153 & 260&373&400 & 31 & 50 & 153 & 260&373&400\\
    \midrule
    \metalstm{}&61.7&75.4&81.4&91.1&72.1&72.7&48.5&{54.5}&{69.7}&74.0&59.1&71.0\\
    
    \midrule
    \bert{} & 75.9 & 84.4 & 88.9 & 94.8 & \textbf{77.5}& \textbf{75.7} & 47.8 & 44.8 & \textbf{75.2} & 82.8&64.0 & 72.9\\
    \mbert{} & \textbf{81.4} & \textbf{86.6} & \textbf{90.0} & \textbf{95.0} & 75.9 & 74.3 & \textbf{64.6} & \textbf{51.1} & 73.6 & \textbf{87.5} & \textbf{64.2} & \textbf{73.8} \\
    \midrule
    \mmetalstm{} &52.9&63.8&65.6&87.6&65.5&61.5&25.6&36.6&42.5&59.2&33.6&46.9\\
    \msmallbert{} &\textbf{76.6}&\textbf{86.0}&\textbf{86.9}&\textbf{95.0}&\textbf{75.4}&\textbf{74.6} &\textbf{59.7}&\textbf{47.6}&\textbf{64.8}&\textbf{81.6}&\textbf{59.4}&\textbf{71.7}\\
    \bottomrule
    \end{tabular}
    \caption{POS tagging and Morphology F1 for all models on low-resource languages. Multilingual models are prefixed with `\textit{m}'.} 
    \label{tab:low_res_transfer}
\end{table*}

\paragraph{Multilingual Modeling Results}
We compare per-language models trained on single language treebanks with multilingual models in Table~\ref{tab:big_mutlilingual_summary} and Table~\ref{tab:mutlilingual_summary}. In the experimental results we use a prefix $m$ to denote the model is a single multilingual model. We compare \metalstm{}, \bert{}, and \smallbert.


\paragraph{Multilingual Models Comparison}
 \mbert{} performs the best among all multilingual models. The smallest and fastest model, \msmallbert{}, performs comparably to \mbert{}, and outperforms \mmetalstm{}, a state-of-the-art model for this task. 

\paragraph{Multilingual Models vs Per-Language Models}
When comparing with per-language models, the multilingual models have lower F1. \newcite{DBLP:journals/corr/abs-1904-02099} shows similar results. \metalstm{}, when trained in a multilingual fashion, has bigger drops than BERT in general. Most of the \metalstm{} drop is due to the character-LSTM, which drops by more than 4 points F1.

\subsection{Low Resource Languages}
We pick languages with fewer than 500 training examples to investigate the performance of low-resource languages: Tamil (ta), Marathi (mr), Belarusian (be),  Lithuanian (lt), Armenian (hy), Kazakh (kk)\footnote{The Universal Dependencies data does not have explicit tuning data for hy and kk.}. Table~\ref{tab:low_res_transfer} shows the performance of the models.

\paragraph{BERT Cross-Lingual Transfer}
  While \newcite{DBLP:journals/corr/abs-1904-09077} shows effective zero-shot crosslingual transfer  from English to other high-resource languages, we show that cross-lingual transfer is even effective on low-resource languages when we train on all languages as \mbert{} is significantly better than \bert{} when we have fewer than 50 examples. In these cases, the \msmallbert{} distilled from the multilingual \mbert{} yields results better than training individual BERT models.  The gains becomes less significant when we have more training data. 

\paragraph{\msmallbert{} Effectiveness}
The multilingual baseline \mmetalstm{} does not do well on low-resource languages. On the contrary, \msmallbert{} performs well and outperforms the state-of-the-art \metalstm{} on the POS tagging task and on four out of size languages of the Morphology task. 

\subsection{Codemixed Input}
We use the Universal Dependencies' Hindi-English codemixed data set~\cite{bhat2017joining} to test the model's ability to label code-mixed data. This dataset is based on code-switching tweets of Hindi and English multilingual speakers. We use the Devanagari script provided by the data set as input tokens.

In the Universal Dependency labeling guidelines, code-switched or foreign-word tokens are labeled as X along with other tokens that cannot be labeled\footnote{https://universaldependencies.org/u/pos/X.html}. The trained model learns to partition the languages in a codemixed input by labeling tokens in one language with X, and tokens in the other language with any of the other POS tags. It turns out that the 2nd-most likely label is usually the correct label in this case; we evaluate on this label when the 1-best is X.


Table \ref{tab:codemixed} shows that all multilingual models handle codemixed data reasonably well without supervised codemixed traininig data.


\begin{table}[t]
\small
    \centering
    \begin{tabular}{lcccccc}
    \toprule
Model & F1 \\
\midrule
\bert{} Supervised& 90.6\\
\metalstm{} English-Only & 47.7 \\
\metalstm{} Hindi-Only & 53.8 \\
\midrule
\mmetalstm& \textbf{83.4}  \\
\mbert &82.9 \\
\msmallbert{} & 79.5 \\
    \bottomrule
    \end{tabular}
    \caption{F1 score on Hindi-English codemixed POS tagging task. Each multilingual model is within 10 points of the supervised BERT model without having explicitly seen code-mixed data.} 
 \label{tab:codemixed}
\end{table}



\section{Conclusion}
We have described the benefits of multilingual models over models trained on a single language for a single task, and have shown that it is possible to resolve a major concern of deploying large BERT-based models by distilling our multilingual model into one that maintains the quality wins with performance fast enough to run on a single CPU. Our distilled model outperforms a multilingual version of a very strong baseline model, and for most languages yields comparable or better performance to a large BERT model.


\bibliography{emnlp-ijcnlp-2019}
\bibliographystyle{acl_natbib}
\newpage

\appendix
\section{Detailed Hyperparameters}
\subsection{Training Hyperparameters}

We use exactly the same hyperparameters as the public multilingual BERT for finetuning our models. We train the part-of-speech tagging task for 10 epochs and the morphology task for 50 epochs.

For distillation, we use the following hyperparameters for all tasks.
\begin{itemize}
    \item learning rate: 1e-4
    \item temperature: 3
    \item batch size: 256
    \item num epochs: 24
\end{itemize}

We take the Wikipedia pretraining data as is and drop sentences with fewer than 10 characters. 

\subsection{Small BERT structure}
We use the vocab and wordpiece model included with the cased public multilingual model on GitHub.

We use the BERT configuration of the public multilingual BERT with the following modifications for \msmallbert{}.
\begin{itemize}
\item Hidden size = 256
\item Intermediate layer size = 1024
\item Num attention heads = 4
\item Layers = 3
\end{itemize}

\section{Detailed Results}
\subsection{The Importance of Distillation}
To understand the importance of distillation in training \msmallbert{}, we compare it to a model with the \smallbert{} structure trained from scratch using only labeled multilingual data the teacher is trained on. Table \ref{tab:distillation_ablation} shows that distillation plays an important role in closing the accuracy gap between teacher and student.

\begin{table}[htb]
\small
    \centering
    \begin{tabular}{lccc}
    \toprule
    Model  & Distilled & Part-of-Speech F1 & Morphology F1
    \\\midrule
    \mbert{} & No & {94.5} & {91.0} \\
    \msmallbert{} & Yes & 93.7 &  88.6\\
    \msmallbert{} & No & 90.2 & 85.5 \\
    \bottomrule
    \end{tabular}
    \caption{Ablation study to show the effect of distillation. The model without distillation has 3.5 points lower macro-averaged F1 on the part-of-speech task and 3.1 lower F1 on the morphology task.} 
    \label{tab:distillation_ablation}
\end{table}

\subsection{Per-Language Results}
We show per-language F1 results of each model in Table~\ref{tab:all_pos} and Table~\ref{tab:all_morph}. For per-language models, no models are trained for treebanks without tuning data, and metrics of those languages are not reported. All macro-averaged results reported exclude those languages.



\onecolumn
\begin{longtable}{lccccc}
\toprule
treebank&\bert{}&\metalstm{}&\mbert{} &\mmetalstm{} &\msmallbert{} \\
\midrule
af\_afribooms&97.62&97.63&97.49&93.16&96.08\\
am\_att&&&3.28&5.6&3.16\\
ar\_padt&90.46&90.55&90.32&89&90.06\\
ar\_pud&&&71.59&68.96&71.06\\
be\_hse&94.81&91.05&95.02&87.59&94.95\\
bg\_btb&99.01&98.77&98.72&96.43&98.19\\
ca\_ancora&98.84&98.62&98.77&97.57&98.45\\
cs\_cac&99.17&99.43&99.3&98.46&98.48\\
cs\_cltt&87.48&87.25&87.67&87.62&87.53\\
cs\_fictree&98.62&98.63&98.25&97.2&97.18\\
cs\_pdt&99.06&99.07&98.99&98.22&98.61\\
cs\_pud&&&97.13&96.53&97\\
da\_ddt&97.59&97.47&97.18&92.36&95.93\\
de\_gsd&94.81&94.17&94.53&91.94&93.82\\
de\_pud&&&88.76&87.42&88.7\\
el\_gdt&97.97&97.4&97.91&94.87&97.16\\
en\_ewt&95.82&95.45&95.2&92.24&94.19\\
en\_gum&96.22&95.02&94.79&92.33&94.24\\
en\_lines&97.22&96.81&95.79&93.96&95.25\\
en\_partut&96.11&95.9&95.02&93.29&94.61\\
es\_ancora&98.87&98.78&98.17&96.27&97.8\\
es\_gsd&93.7&93.9&89.65&90.61&89.58\\
es\_pud&&&85.87&86.1&85.71\\
et\_edt&97.27&97.17&97.02&94.32&95.64\\
eu\_bdt&96.2&96.1&95.51&91.53&94.15\\
fa\_seraji&97.57&97.17&97.17&95.29&96.92\\
fi\_ftb&96.26&96.12&93.15&87.23&89.79\\
fi\_pud&&&95.55&93.23&95.01\\
fi\_tdt&96.81&97.02&93.9&91.58&92.6\\
fr\_gsd&96.62&96.45&96.23&95.37&96.05\\
fr\_partut&96.18&96&95.43&94.35&94.93\\
fr\_pud&&&90.77&90.1&90.64\\
fr\_sequoia&96.77&97.59&97.07&95.91&96.75\\
fr\_spoken&97.55&95.78&96.1&90.07&93.25\\
ga\_idt&91.92&91.55&90.83&84.16&85.72\\
gl\_ctg&96.99&97.21&96.5&92.87&95.84\\
gl\_treegal&&&93.4&91.28&91.9\\
he\_htb&82.76&82.49&82.69&80.93&81.93\\
hi\_hdtb&97.31&97.39&97.1&96.2&96.43\\
hi\_pud&&&86.48&85.33&85.68\\
hr\_set&97.79&97.94&97.47&96.24&97.2\\
hu\_szeged&96.51&94.71&95.99&85.5&95.47\\
hy\_armtdp&84.42&&86.62&63.82&86.98\\
id\_gsd&93.06&93.37&93.3&90.81&93.35\\
id\_pud&&&63.52&63.5&63.33\\
it\_isdt&98.33&98.06&98.27&96.7&97.8\\
it\_partut&98.12&98.17&98.09&96.99&98.06\\
it\_postwita&95.66&95.86&95.6&94.17&93.2\\
it\_pud&&&93.84&92.72&93.67\\
ja\_gsd&88.63&88.73&88.54&87.03&88.43\\
ja\_modern&&&41.55&51.26&21.61\\
ja\_pud&&&89.15&87.96&89.3\\
kk\_ktb&75.93&61.7&81.36&52.91&80.06\\
ko\_gsd&95.92&95.64&90.3&86.39&88.62\\
ko\_kaist&95.56&95.42&93.86&87.46&93.43\\
ko\_pud&&&41.93&46.11&31.96\\
la\_ittb&98.34&98.42&98.3&97.18&97.65\\
la\_perseus&&&89.91&83.85&85.23\\
la\_proiel&96.34&96.37&95.97&92.02&93.78\\
lt\_hse&88.88&81.43&90.01&65.6&86.9\\
lv\_lvtb&94.79&94.47&93.71&88.25&91.3\\
mr\_ufal&77.45&72.1&75.92&65.48&75.41\\
nl\_alpino&97.1&96.16&97.33&93.78&96.19\\
nl\_lassysmall&95.54&95.92&95.72&94.4&95.47\\
no\_bokmaal&98&98&97.95&95.27&97.04\\
no\_nynorsklia&&&94.08&88.27&92.55\\
no\_nynorsk&97.94&97.92&97.69&94.91&96.59\\
pl\_lfg&98.7&98.5&98.39&95.21&97.48\\
pl\_sz&98.56&97.91&98.05&94.73&97.29\\
pt\_bosque&96.74&96.73&96.16&95.53&95.85\\
pt\_gsd&95.83&95.44&93.84&93.07&94.44\\
pt\_pud&&&89.48&89.66&89.29\\
ro\_nonstandard&94.67&94.48&94&92.05&91.9\\
ro\_rrt&97.63&97.52&97.47&95.78&96.71\\
ru\_gsd&92.23&91.39&90.84&88.13&90.14\\
ru\_pud&&&89.7&88.92&89.52\\
ru\_syntagrus&98.3&98.65&98.32&97.13&98.03\\
ru\_taiga&&&93.62&92.75&93.18\\
sa\_ufal&&&32.47&29.58&27.11\\
sk\_snk&97.08&96.32&96.98&93.61&96.35\\
sl\_ssj&97.07&96.68&96.89&94.24&95.58\\
sl\_sst&&&94.51&90.34&91.79\\
sr\_set&98.63&98.33&98.31&94.79&97.36\\
sv\_lines&97.21&96.59&96.99&93.64&95.57\\
sv\_pud&&&94.52&92.06&94.32\\
sv\_talbanken&98.03&97.34&97.77&94.91&96.76\\
ta\_ttb&75.71&72.7&74.28&61.51&74.6\\
te\_mtg&94.25&92.72&93.42&87.32&93.42\\
th\_pud&&&2.37&2.73&1.54\\
tl\_trg&&&70.69&28.62&68.28\\
tr\_imst&93.96&94.03&93.1&84.64&91.8\\
tr\_pud&&&73.1&68.36&72.47\\
uk\_iu&97.29&96.6&97.28&93&96.88\\
ur\_udtb&93.83&93.87&93.69&93&93.05\\
vi\_vtb&77.67&76.42&77.44&72.01&77.06\\
yo\_ytb&&&43.48&30.85&34.59\\
zh\_cfl&&&49.83&39.77&49.42\\
zh\_gsd&87.6&85.7&85.96&82.76&86.08\\
zh\_hk&&&66.29&57.88&65.86\\
zh\_pud&&&83.3&73.3&82.95\\
\bottomrule
\caption{POS tagging F1 of all models.}
\label{tab:all_pos}
\end{longtable}

\begin{longtable}{lccccc}
\toprule
treebank&\bert{} F1&\metalstm{} F1&\mbert{} F1&\mmetalstm{} F1&\msmallbert{} F1\\
\midrule
af\_afribooms&97.11&97.36&96.53&88.98&93.75\\
am\_att&&&&32.36&32.36\\
ar\_padt&88.26&88.24&87.76&83.14&85.34\\
ar\_pud&&&36.33&34.28&36.08\\
be\_hse&82.83&74.03&87.52&59.16&81.82\\
bg\_btb&97.54&97.58&97.47&91.41&95.4\\
ca\_ancora&98.37&98.21&98.28&96.04&97.67\\
cs\_cac&96.33&96.49&96.54&88.11&93.47\\
cs\_cltt&81.61&79.89&83.86&78.82&80.61\\
cs\_fictree&96.39&96.4&94.09&83.37&87.59\\
cs\_pdt&97.18&96.91&97.15&89.77&94.63\\
cs\_pud&&&93.88&87.44&91.81\\
da\_ddt&97.22&97.08&95.62&89.82&94.08\\
de\_gsd&90.84&90.58&90.4&80.69&88.99\\
de\_pud&&&30.41&30.55&30.4\\
el\_gdt&94.57&93.95&94.83&87.6&92.07\\
en\_gum&96.87&96&93.79&90.11&93.71\\
en\_lines&97.32&96.68&93.11&87.49&92.07\\
en\_partut&94.88&95.38&90.76&79.99&90.18\\
en\_pud&&&93.25&91.23&93.1\\
es\_ancora&98.45&98.42&97.6&95.17&97\\
es\_gsd&93.52&93.72&88.72&89.26&88.78\\
es\_pud&&&52.7&52.8&52.73\\
et\_edt&96.14&96.11&95.78&90.51&92.14\\
eu\_bdt&93.27&92.56&92.67&76.72&84.53\\
fa\_seraji&97.35&97.25&96.91&93.82&96.28\\
fi\_ftb&96.34&96.48&92.32&77.89&86.47\\
fi\_pud&&&93.58&91.12&91.65\\
fi\_tdt&95.03&95.58&90.96&88.44&87.48\\
fr\_gsd&96.05&96.11&94.67&86.97&94.51\\
fr\_partut&93.32&92.93&88.9&87.48&87.05\\
fr\_pud&&&59.15&57.5&58.94\\
fr\_sequoia&97.09&97.13&91.54&85.23&90.74\\
fr\_spoken&100&100&98.62&80.67&96.67\\
ga\_idt&82.2&81.78&81.2&63.44&66.82\\
gl\_ctg&98.98&98.95&95.27&89.98&95.1\\
gl\_treegal&&&80.05&68.73&75.97\\
he\_htb&81.27&80.85&80.79&76.89&78.74\\
hi\_hdtb&93.32&93.85&92.91&89.09&90.65\\
hi\_pud&&&22.1&22.37&22.03\\
hr\_set&91.99&91.85&91.24&81.62&87.81\\
hu\_szeged&93.65&91.28&92.93&71.25&87.36\\
hy\_armtdp&41.13&54.45&51.08&36.59&46.43\\
id\_gsd&94.84&96&94.85&91.62&94.39\\
id\_pud&&&39.83&42.79&39.79\\
it\_isdt&97.7&97.82&97.87&95.47&97.37\\
it\_partut&97.35&97.73&98.01&96.33&97.9\\
it\_postwita&95.62&96.05&95.03&91.52&93.17\\
it\_pud&&&57.82&57.41&57.6\\
ja\_gsd&90.29&90.45&90.29&90.39&90.41\\
ja\_modern&&&63.9&61.17&63.99\\
ja\_pud&&&57.4&57.26&57.27\\
kk\_ktb&&&64.6&25.55&59.49\\
ko\_gsd&99.62&99.55&99.4&98.99&99.37\\
ko\_kaist&100&100&99.94&99.24&99.93\\
ko\_pud&&&38.33&38.66&38.27\\
la\_ittb&96.7&96.94&97.15&90.78&93.91\\
la\_perseus&&&82.09&64.73&72.24\\
la\_proiel&90.82&91.01&91.51&79.08&83.99\\
lt\_hse&75.21&69.65&73.61&42.51&65.22\\
lv\_lvtb&88.61&91.34&88.1&79.11&81.91\\
mr\_ufal&63.95&59.11&64.2&33.63&54.01\\
nl\_alpino&96.22&96.13&96.53&91.9&95.67\\
nl\_lassysmall&96.46&96.02&95.55&92.16&95.28\\
no\_bokmaal&96.85&97.13&96.48&91.17&95.31\\
no\_nynorsklia&&&94.22&89.56&91.08\\
no\_nynorsk&96.7&97.04&96.49&92.12&94.79\\
pl\_lfg&95.85&94.68&84.96&47.99&84.56\\
pl\_sz&93.9&91.93&71.4&73.02&65.36\\
pt\_bosque&96.27&96.16&87.04&83.13&85.72\\
pt\_gsd&97.2&95.33&67.72&76.01&71.88\\
pt\_pud&&&52.06&49.79&50.95\\
ro\_nonstandard&88.52&88.91&86.89&82.1&82.14\\
ro\_rrt&97.02&97.23&96.58&93.2&94.85\\
ru\_gsd&88.83&86.73&81.44&64.2&78.93\\
ru\_pud&&&37.97&35.26&37.49\\
ru\_syntagrus&97.02&96.9&95.99&91.96&94.33\\
ru\_taiga&&&88.56&84.02&86.01\\
sa\_ufal&&&15.9&16.14&16.33\\
sk\_snk&92.06&89.63&91.58&68.25&85.29\\
sl\_ssj&94.39&93.78&94.41&82.69&89.23\\
sl\_sst&88.46&&91.89&78.22&85.59\\
sr\_set&94.83&94.71&92.79&73.51&90.48\\
sv\_lines&89.54&89.55&88.66&83.27&86.4\\
sv\_pud&&&77.39&73.94&76.79\\
sv\_talbanken&96.92&96.56&96.13&90.23&94.49\\
ta\_ttb&72.91&71.01&73.75&46.9&70.22\\
te\_mtg&98.96&98.96&98.54&98.68&98.54\\
th\_pud&&&8.27&0&8.43\\
tl\_trg&&&29.31&28.62&25.17\\
tr\_imst&89.5&91&88.63&73.23&81.99\\
tr\_pud&&&23.72&23.84&23.46\\
uk\_iu&92.4&90.98&92.64&79.49&88.79\\
ur\_udtb&82.24&83.72&82.64&81.89&82.48\\
vi\_vtb&83.74&84&83.93&83.58&83.94\\
yo\_ytb&&&58.78&86.82&61.88\\
zh\_cfl&&&46.55&43.55&45.73\\
zh\_gsd&87.64&88.38&88.31&87.05&88.5\\
zh\_hk&&&66.33&64.97&66.23\\
zh\_pud&&&86.35&83.6&86.14\\
\bottomrule
\caption{Morphology F1 of all models.}
\label{tab:all_morph}
\end{longtable}


\end{document}